\documentclass{article}

\usepackage{booktabs}
\usepackage{caption}
\usepackage{enumitem}
\usepackage{graphicx}
\usepackage{hyperref}
\usepackage{microtype}
\usepackage[binary-units=true]{siunitx}
\usepackage{subfigure}
\usepackage{xcolor}

\usepackage[accepted]{sysml2019}

\sysmltitlerunning{Efficient Memory Management for Deep Neural Net Inference}

\hypersetup{draft}

\begin{document}

\twocolumn[
\sysmltitle{Efficient Memory Management\\for Deep Neural Net Inference}
\begin{sysmlauthorlist}
\sysmlauthor{Yury Pisarchyk}{ggl}
\sysmlauthor{Juhyun Lee}{ggl}
\end{sysmlauthorlist}
\sysmlaffiliation{ggl}{Google Research, Mountain View, CA, USA}
\sysmlcorrespondingauthor{Yury Pisarchyk}{ra16bit@gmail.com}
\sysmlcorrespondingauthor{Juhyun Lee}{impjdi@google.com}
\sysmlkeywords{Neural Network, Inference, Memory}
\vskip 0.25in
\begin{abstract}
While deep neural net inference was considered a task for servers only, latest
advances in technology allow the task of inference to be moved to mobile and
embedded devices, desired for various reasons ranging from latency to privacy.
These devices are not only limited by their compute power and battery, but also
by their inferior physical memory and cache, and thus, an efficient memory
manager becomes a crucial component for deep neural net inference at the edge.
We explore various strategies to smartly share memory buffers
among intermediate tensors in deep neural nets. Employing these can result in
up to 11\% smaller memory footprint than the state of the art.
\end{abstract}
]
\printAffiliationsAndNotice{}

\section{Introduction}
\label{sec:intro}

Deep neural networks are widely used to solve various machine learning problems
including, but not limited to, computer vision, natural language processing,
signal processing, and others. While employing deep neural networks is
technically challenging for its demanding resources in computation and memory,
recent advances in computing hardware enabled deep neural nets to be carried out
on mobile and embedded devices~\cite{lee2019device,wu2019machine}.

Deep neural networks can be represented as directed acyclic graphs (DAG) with
the nodes describing the computational operations such as \textsc{convolution}
or \textsc{softmax} and the edges describing the tensors containing the
intermediate computation results between the operators
\cite{bergstra2010theano}.  These tensors are materialized with memory buffers:
A tensor of shape $[B,H,W,C]$ translates to a memory buffer of size
${B}\times{H}\times{W}\times{C}\times\textrm{sizeof(float)}$.  To reduce the
overhead of dynamic memory allocation, the memory buffers for intermediate
tensors are typically allocated before running the model, but these can take up
a significant amount of memory.  For example, the intermediate tensors of
Inception v3~\cite{szegedy2016rethinking} take up 37\% of \SI{147}{\mega\byte}
total run-time memory and those of MobileNet v2 \cite{sandler2018mobilenetv2}
consume 63\% of \SI{41}{\mega\byte}.

Fortunately, the intermediate tensors do not have to co-exist in memory; thanks
to the mostly sequential execution of the network due to data dependency, only
one operator is active at any given point in time, and only its immediate input
and output intermediate tensors are needed.  Thus, we explore the idea of
reusing the memory buffers to optimize the total memory footprint of the deep
neural net inference engine.  If the DAG has the shape of a simple chain, memory
buffers can be reused in alternating fashion, assuming the memory buffers have
enough capacity to contain any intermediate tensor in the network.  However, the
reusing problem is not trivial to solve if memory buffers have limited capacity
or the network contains residual connections~\cite{he2016deep}.

In this paper, we present five strategies for efficient memory sharing for
intermediate tensors.  They show up to $10.5\times$ reduction compared to
keeping all of the intermediate tensors na\"ively in memory and up to 11\%
reduction of memory consumption compared to prior state of the art.  Efficiently reusing
memory buffers leads to improved cache hit rate that can also translate to up to
10\% improvement in inference speed.  These strategies are applicable to neural
net inference only and not to training as intermediate tensors need to be kept
alive and thus their memory cannot be re-purposed.

\section{Related Work}

Efficiently managing memory for deep neural networks is not only a problem for
resource-constrained environment
% such as mobile or embedded devices; it has been a challenge even
but also for servers.  MXNet employs a number of techniques for reducing
memory consumption such as in-place operators and intermediate tensors memory
co-share, using simple heuristic algorithm for memory allocation, that is safe
for parallel operators execution~\cite{chen2015mxnet}.  However, the authors do
not focus on the core problem of memory management and do not explore different
algorithms that can solve this problem in the most effective way.
\cite{chen2016training} employs a similar technique along with trading
computation for memory, but is not suitable for mobile.

Caffe2's on-device inference engine employs NNPACK and QNNPACK
\cite{wu2019machine}.  These neural network operator libraries choose the best
data layout and optimize off-chip memory access, but do not focus on minimizing
the inference engine's memory footprint.  \cite{li2016optimizing} similarly
exploits data layout to optimize off-chip memory access and focuses on
server-like settings which are not as resource-constrained as mobile or embedded
devices.

\begin{figure}[!t]  % Figure 1.  Page 2 top left.
  \centering
    \includegraphics[height=4.5cm]{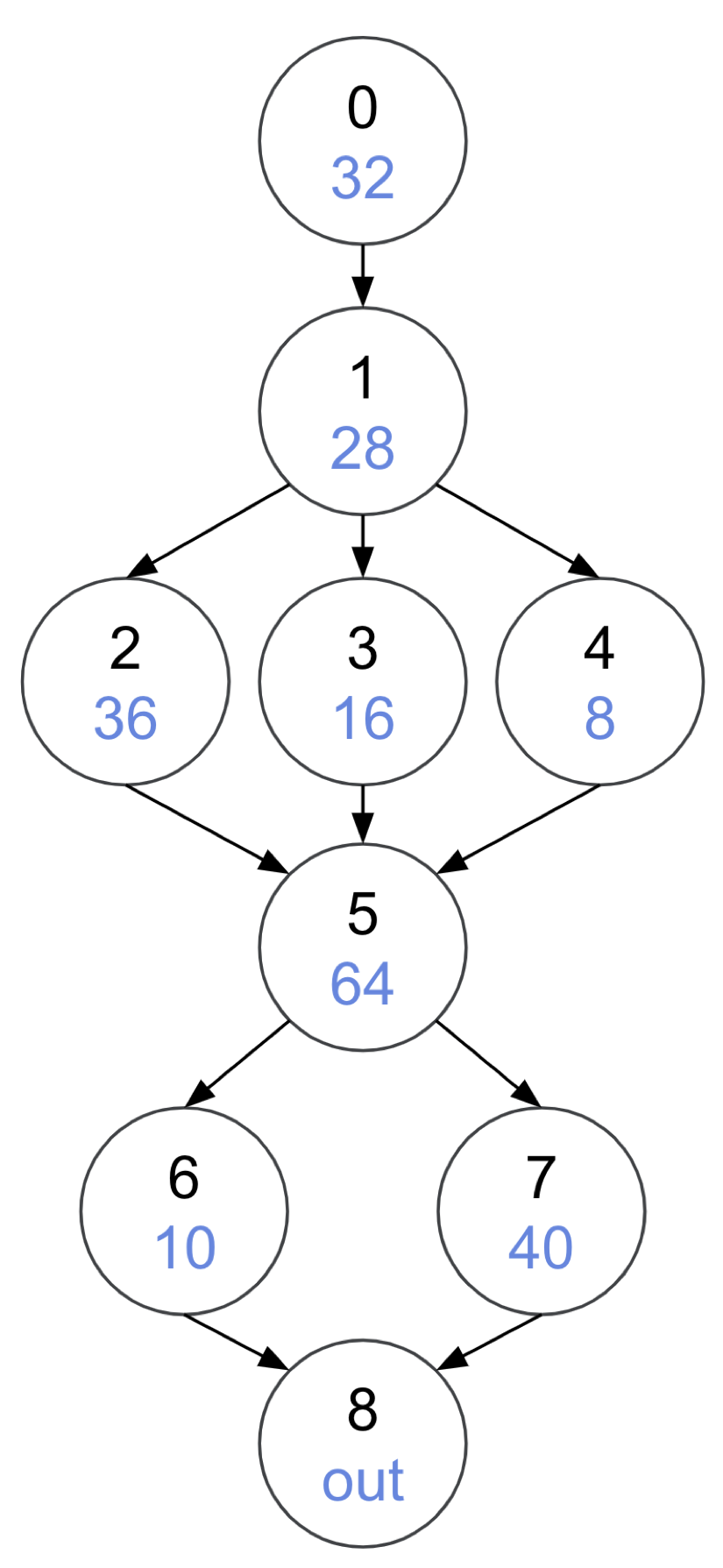}
    \hspace{0.1\textwidth}
    \includegraphics[height=4.5cm]{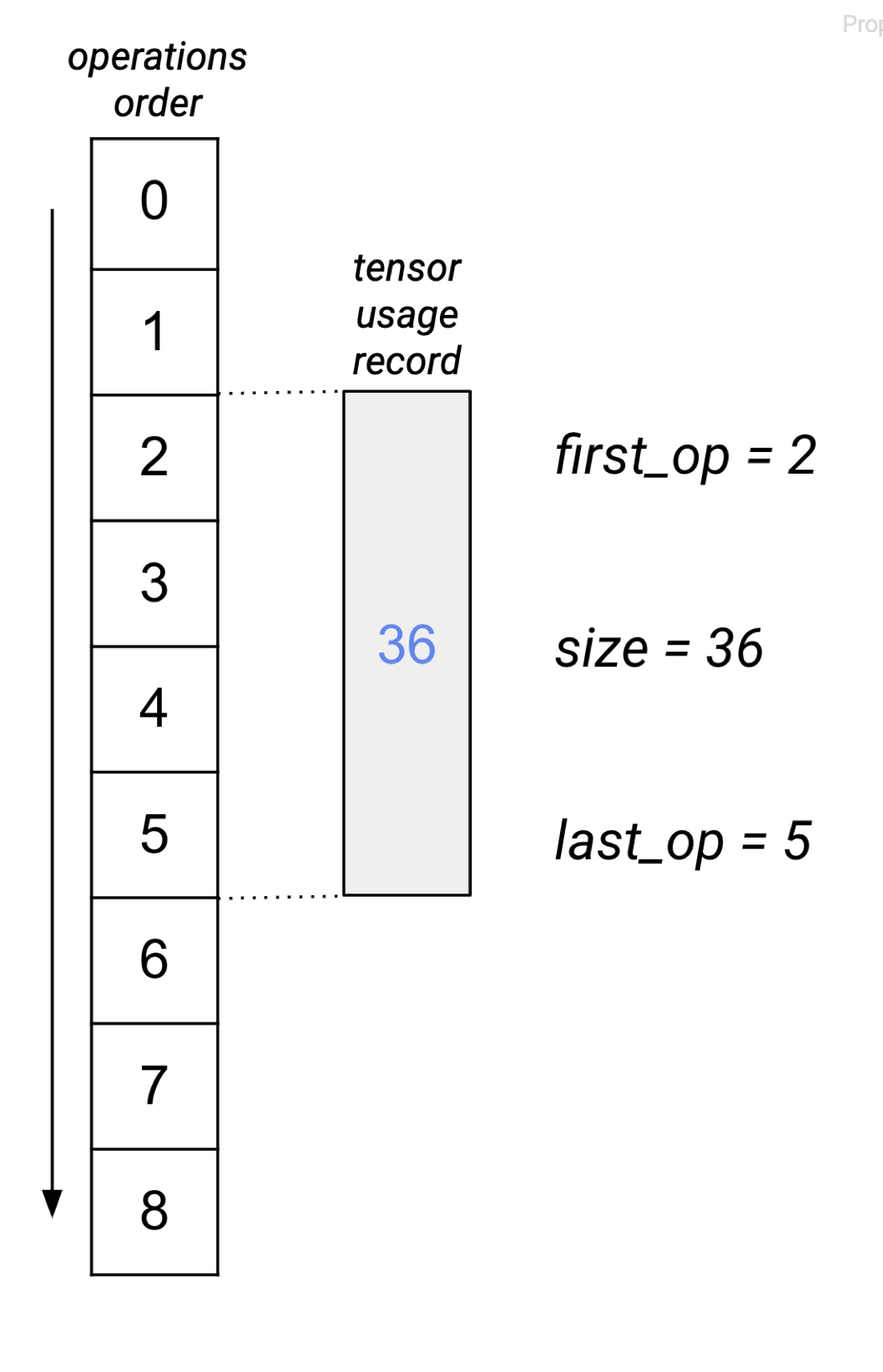}
  \caption{
    (a) An example neural network with each of its intermediate tensors denoted
    with its index and its size in black and blue, respectively; note that
    tensor \#8 is not an intermediate tensor.
    (b) The tensor usage record of tensor \#2 as an example.}
    \vspace{-0.3cm}
  \label{fig:example}
\end{figure}

TensorFlow Lite (TFLite) GPU employs a memory manager for its
GPU buffers~\cite{lee2019device}.  Two approximations are explored for
this NP-complete resource management problem~\cite{sethi1975complete},
which is similar to register allocation problem, but more complex due to different sizes of tensors.
\cite{sekiyama2018profile} solve the memory allocation problem as a special
case of 2D strip packing problem.  We present strategies that
outperform these in most cases.

\section{Definition of Terms}

% A memory buffer in use by a tensor can be reused by another tensor (``shared'')
% if the former tensor's last consumer finished executing.  For an efficient
% memory manager, it is crucial to analyze and understand this data dependency.
This section defines several key terms that should facilitate strategy
description in the following sections.

\textbf{Tensor Usage Interval} of an intermediate tensor $t$ is defined as the
pair $\{\mathit{first\_op_t}, \mathit{last\_op_t}\}$, where 
$\mathit{first\_op_t}$ and $\mathit{last\_op_t}$ are the indices of the first
and the last operator that use $t$ as its input or output, respectively.
The indices are from a topological sort of the neural network which is also
the operators' execution order.  For the remainder of the paper,
we assume that this order is fixed.  Note that no two
tensors with intersecting  usage intervals can share memory.

\begin{figure}[!t]  % Figure 2.  Page 2 top right.
  \centering
  \includegraphics[height=4.5cm]{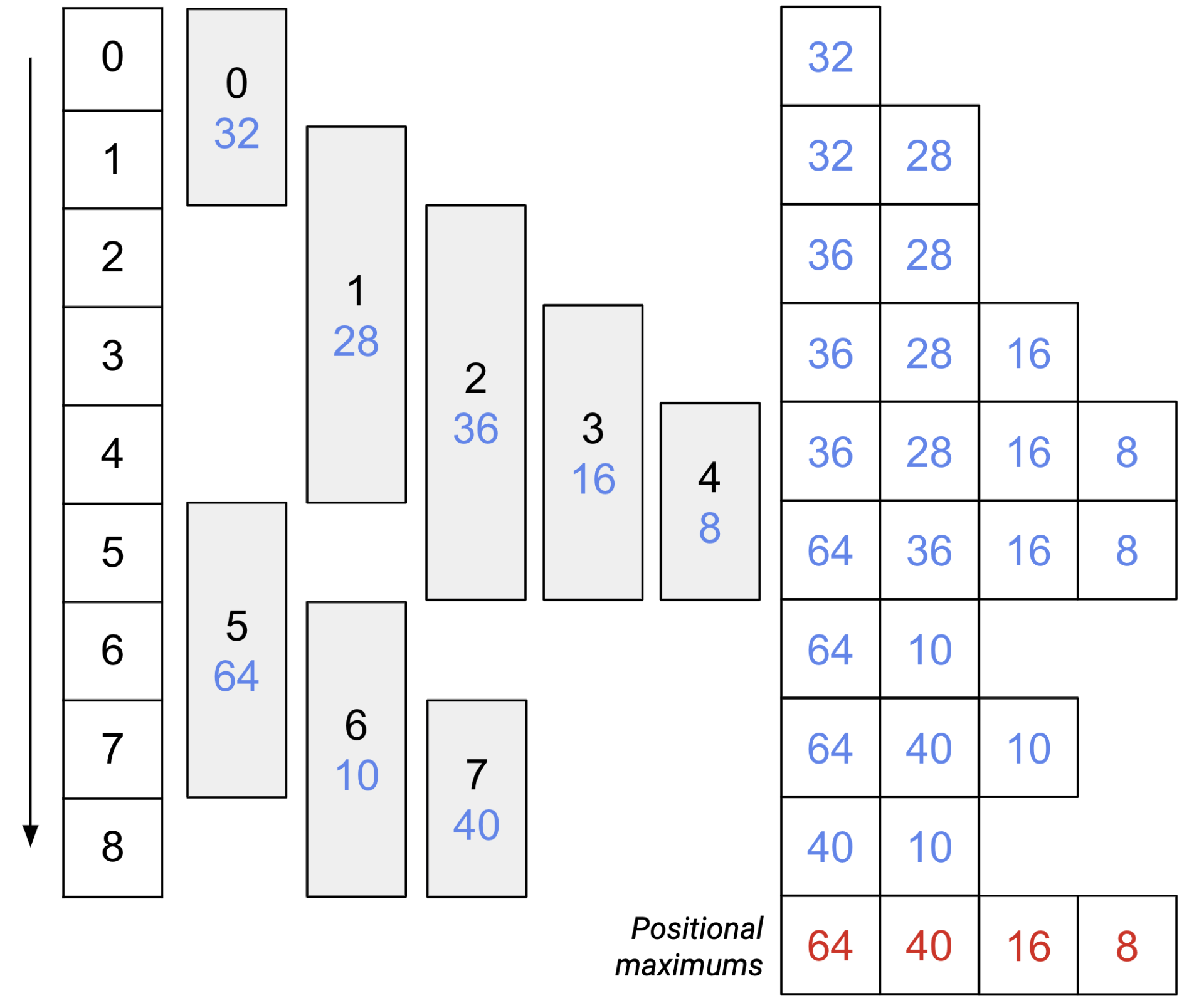}
  \caption{
    (a) All tensor usage records for the graph in Figure~\ref{fig:example} (a).
    (b) The sorted operator profiles at each execution timestamp; only the
    tensor sizes are shown in blue.  The positional maximums of each column are
    on the final row, denoted in red.
  }
  \vspace{-0.3cm}
  \label{fig:profiles}
\end{figure}

\textbf{Tensor Usage Record} of an intermediate tensor $t$ is defined as the
triple $\{\mathit{first\_op_t}, \mathit{last\_op_t}, \mathit{size_t}\}$, where
$\mathit{size_t}$ is $t$'s aligned size in bytes.  Figure~\ref{fig:example}
illustrates an example neural network and the tensor usage record of tensor \#2.
The full set of tensor usage records is depicted in Figure~\ref{fig:profiles} (a).

\textbf{Operator Profile} of an operator $op$ is defined as the set of all
tensor usage records $t$ such that $op$ falls between
$\mathit{first\_op_t}$ and $\mathit{last\_op_t}$.  Figure~\ref{fig:profiles} (b)
visualizes the operator profile of each operator sorted in descending order by
size.

\textbf{Operator Breadth} of an operator is defined as the sum of all tensor
sizes in its profile.  For example, operator \#3 in Figure \ref{fig:profiles}
(b) has the operator breadth of $36 + 28 + 16 = 80$.

The $i$-th \textbf{Positional Maximum} is the maximum across $i$-th tensor sizes
in descending order for each operator profile.  For example, the third
positional maximum in Figure~\ref{fig:profiles} (b) is equal to
$\mathit{max(16, 16, 16, 10)} = 16$.
% The sum of the
% positional maximums is the theoretical lower bound for the Shared Objects problem
% (Section~\ref{sec:shared-objects}).

\section{The Shared Objects Approach}
\label{sec:shared-objects}

There are broadly two ways of sharing memory which are discussed in this and the
following section.  We call the first \textit{Shared Objects} where each
memory buffer (``shared object'') is assigned to an intermediate tensor at a
given time.  No two tensors with intersecting usage intervals can be assigned to
the same shared object and no shared object can be used for more than one tensor
at any moment in time.  The size of the shared object is the maximum of all the
tensor sizes it is assigned to.  The main objective is \textit{to minimize the
total size of these shared objects}.  This approach is most suitable for GPU
textures.
% To get a baseline, we first investigate the theoretical lower bound for this problem.

% For each operator of the neural network there is a subset of tensors, that must be already allocated before its execution. We call this subset an operator \textit{profile}. The profile for operator $op_i$ consists only of those tensors $t$, for which $op_i$ lies inside of their usage interval, e.g. $first\_op_t \leq op_i \leq last\_op_t$. By operator $op_i$ \textit{breadth} we mean minimum possible size of allocated memory during its execution. It is equal to the sum of tensor sizes in its profile.
% Calculation of operators breadths and sorted profiles is summarized in Algorithm
% \ref{alg:profile-and-breadth}.

\subsection{Theoretical Lower Bound}

Each operator profile is sorted in non-increasing order by its tensor sizes.
The largest shared object in the resulting allocation will have a size greater
or equal to the largest of first elements across all sorted profiles, and the
second largest shared object cannot be less in size than largest of second
elements across sorted profiles.  This property holds for every shared object.
The number of shared objects cannot be less than the largest number of tensors
in one operator profile.  Thus, the sum of the positional maximums is the
theoretical lower bound for the Shared Objects problem.
This lower bound may not be achievable for some neural networks.

\subsection{Greedy by Breadth}

% As we can note from the investigation on the lower bound,
Operator breadths are more correlated for the resulting memory consumption than
the order of tensor allocations during inference.  Thus, we start from the
allocation of tensors that must be present in memory during execution of 
operator with greater breadth, i.e.~\textit{Greedy by Breadth} (Algorithm
\ref{alg:greedy-by-breadth}).

\begin{algorithm}[!b]
  \begin{small}
   \caption{Greedy by Breadth for Shared Objects}
   \label{alg:greedy-by-breadth}
\begin{algorithmic}[1]
   \STATE $\mathit{shared\_objects} \leftarrow \emptyset$
   \STATE \textbf{for each} $\mathit{t} \in$ tensor~usage~records \textbf{do}
   \STATE \hspace{0.2cm}$\mathit{assigned\_shared\_object_t} \leftarrow \mathit{NIL}$
   \STATE sort $\mathit{operators}$ in non-increasing order of $\mathit{breadth}$
   \STATE \textbf{for each} $op \in \mathit{operators}$ \textbf{do}
   \STATE \hspace{0.2cm}\textbf{for each} $\mathit{t} \in \mathit{profile_{op}}$ \textbf{do}
   \STATE \hspace{0.4cm} \textbf{if} $\mathit{assigned\_shared\_object_t}\neq\mathit{NIL}$  \textbf{then}
   \STATE \hspace{0.6cm} \textbf{continue}
   \STATE \hspace{0.4cm} $\mathit{best\_obj} \leftarrow \mathit{NIL}$
   \STATE \hspace{0.4cm} \textbf{for each} $obj \in \mathit{shared\_objects}$ \textbf{do}
   \STATE \hspace{0.6cm} $\mathit{is\_better} \leftarrow \mathit{TRUE}$
   \STATE \hspace{0.6cm} \textbf{if} $\mathit{best\_obj} \neq \mathit{NIL}$ \textbf{then}
   \STATE \hspace{0.8cm} \textbf{if} $\mathit{best\_obj.size} < \mathit{size_t}$
   \STATE \hspace{1.0cm} \textbf{if} $\mathit{obj.size} \le \mathit{best\_obj.size}$ \textbf{then}
   \STATE \hspace{1.2cm} $\mathit{is\_better} \leftarrow \mathit{FALSE}$
   \STATE \hspace{0.8cm} \textbf{elseif} $\mathit{obj.size}\ge\mathit{best\_obj.size}$ \textbf{or} \\
          \hspace{1.0cm} $\mathit{obj.size} < \mathit{size_t}$ \textbf{then}
   \STATE \hspace{1.0cm} $\mathit{is\_better} \leftarrow \mathit{FALSE}$
   \STATE \hspace{0.6cm} $\mathit{suitable} \leftarrow \mathit{TRUE}$
   \STATE \hspace{0.6cm} \textbf{for each} $\mathit{x} \in$ tensor~usage~records \textbf{do}
   \STATE \hspace{0.8cm} $\mathit{max\_first\_op}\leftarrow\max(\mathit{first\_op_t},\mathit{first\_op_x})$
   \STATE \hspace{0.8cm} $\mathit{min\_last\_op}\leftarrow\min(\mathit{last\_op_t},\mathit{last\_op_x})$
   \STATE \hspace{0.8cm} \textbf{if} $\mathit{assigned\_shared\_object_x}=\mathit{obj}$ \textbf{and} \\
          \hspace{1.0cm}$max\_first\_op\le min\_last\_op$ \textbf{then}
   \STATE \hspace{1.0cm} $\mathit{suitable} \leftarrow \mathit{FALSE}$
   \STATE \hspace{0.6cm} \textbf{if} $\mathit{suitable}$ \textbf{and} $\mathit{is\_better}$ \textbf{then}
   \STATE \hspace{0.8cm} $\mathit{best\_obj} \leftarrow \mathit{obj}$
   \STATE \hspace{0.4cm} \textbf{if} $\mathit{best\_obj} = \mathit{NIL}$ \textbf{then}
   \STATE \hspace{0.6cm} $\mathit{best\_obj} \leftarrow$ new shared object with size $\mathit{size_t}$
   \STATE \hspace{0.6cm} $\mathit{shared\_objects}$.insert($\mathit{best\_obj}$)
   \STATE \hspace{0.4cm} \textbf{else}
   \STATE \hspace{0.6cm} $\mathit{best\_obj.size} \leftarrow \max(\mathit{best\_obj.size}, \mathit{size_t})$
   \STATE \hspace{0.4cm} $\mathit{assigned\_shared\_object_t} \leftarrow \mathit{best\_obj}$
\end{algorithmic}
\end{small}
\end{algorithm}

Operators are sorted in non-increasing order by their breadth (L.4).  For each operator in this sorted ordering, we assign shared objects to tensors from its profile, but only for those that have not been assigned yet (L.7). If there are several such tensors, we start from the largest by size.  A shared object $s$ is \textit{suitable} for assignment to tensor $t$, if and only if there is no tensor $u$, such that $s$ is assigned to the $u$ and usage intervals of $t$ and $u$ overlap (L.18--23).
Shared object assignment (L.12--17, 24--28) can be summarized as:
\begin{itemize}[noitemsep,topsep=0pt,leftmargin=*]
\item If there are suitable shared objects not smaller than $size_t$, assign the smallest to $t$.
\item If all suitable shared objects are smaller than $size_t$, update the largest size to $size_t$ and assign it to $t$.
\item If there are no suitable shared objects, create a new shared object with size $size_t$ and assign it to $t$.
\end{itemize}

\begin{figure}[!t]  % Figure 3.  Page 3 upper right.
  \centering
  \includegraphics[height=4.5cm]{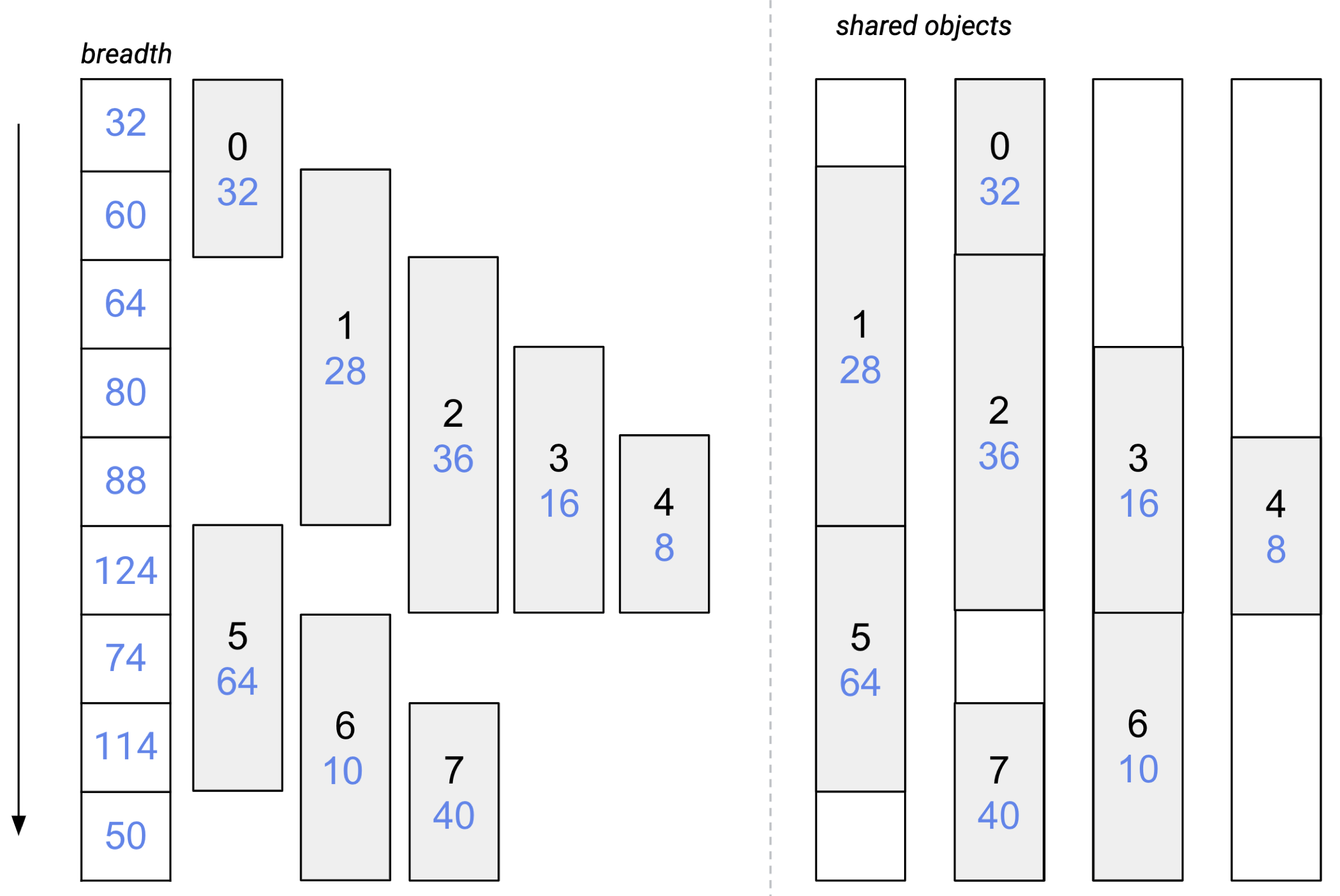}
  \caption{Greedy by Breadth for Shared Objects on the neural network in Figure~\ref{fig:example}.
  % To the left operators' breadths are shown in blue.
  }
  \vspace{-0.2cm}
  \label{fig:greedy-by-breadth}
\end{figure}

The algorithm has a run-time complexity of $O(kn^2)$, where $k$ and $n$ are the
number of shared objects and intermediate tensors, respectively,
when implemented na\"ively.  Note that $k$ is often at lower tens, whereby $n$
is one or two magnitudes larger in a typical neural network.  With an interval
tree for each shared object that stores the usage intervals of all tensors, the
complexity can be reduced to $O(kn\log n)$.  An example output is shown in
Figure~\ref{fig:greedy-by-breadth}.

\subsection{Greedy by Size}

\begin{algorithm}[!b]
\begin{small}
   \caption{Greedy by Size for Shared Objects}
   \label{alg:greedy-by-size}
\begin{algorithmic}[1]
   \STATE sort tensor~usage~records in non-increasing order of $\mathit{size}$
   \STATE $\mathit{shared\_objects} \leftarrow \emptyset$
   \STATE \textbf{for each} $\mathit{t} \in$ tensor~usage~records \textbf{do}
   \STATE \hspace{0.2cm}$\mathit{assigned\_shared\_object_t} \leftarrow \mathit{NIL}$
   \STATE \textbf{for each} $\mathit{t} \in$ tensor~usage~records \textbf{do}
   \STATE \hspace{0.2cm} $\mathit{best\_obj} \leftarrow \mathit{NIL}$
   \STATE \hspace{0.2cm} \textbf{for each} $obj \in \mathit{shared\_objects}$ \textbf{do}
   \STATE \hspace{0.4cm} $\mathit{suitable} \leftarrow \mathit{TRUE}$
   \STATE \hspace{0.4cm} \textbf{for each} $\mathit{x} \in$ tensor~usage~records \textbf{do}
   \STATE \hspace{0.6cm} $\mathit{max\_first\_op}\leftarrow\max(\mathit{first\_op_t},\mathit{first\_op_x})$
   \STATE \hspace{0.6cm} $\mathit{min\_last\_op}\leftarrow\min(\mathit{last\_op_t},\mathit{last\_op_x})$
   \STATE \hspace{0.6cm} \textbf{if} $\mathit{assigned\_shared\_object_x}=\mathit{obj}$ \textbf{and} \\
   \hspace{0.8cm} $\mathit{max\_first\_op}\le\mathit{min\_last\_op}$ \textbf{then}
   \STATE \hspace{0.8cm} $\mathit{suitable} \leftarrow \mathit{FALSE}$
   \STATE \hspace{0.4cm} \textbf{if} $\mathit{suitable}$ \textbf{then}
   \STATE \hspace{0.6cm} $\mathit{best\_obj} \leftarrow \mathit{obj}$
   \STATE \hspace{0.2cm} \textbf{if} $\mathit{best\_obj} = \mathit{NIL}$ \textbf{then}
   \STATE \hspace{0.4cm} $\mathit{best\_obj} \leftarrow$ new shared object with size $\mathit{size_t}$
   \STATE \hspace{0.4cm} $\mathit{shared\_objects}$.insert($\mathit{best\_obj}$)
   \STATE \hspace{0.2cm} $\mathit{assigned\_shared\_object_t} \leftarrow \mathit{best\_obj}$
\end{algorithmic}
\end{small}
\end{algorithm}

While the operator breadth is significant, it is a number derived from
tensor sizes in the operator profiles.  Thus, we explore another strategy
\textit{Greedy by Size} where the tensor sizes are the most significant feature
(Algorithm~\ref{alg:greedy-by-size}).
% and larger tensors are preferred over smaller tensors

We iterate through intermediate tensors in non-increasing order of their size (L.1,5), and for each tensor $t$ find \textit{suitable} shared object to assign to it (L.8--11), similar to Greedy by Breadth. As before, there are no shared objects in the beginning (L.2).
The assignment becomes even easier, because we prefer larger tensors over smaller ones, i.e.~the shared object size never increases, and only two steps remain:
\begin{itemize}[noitemsep,topsep=0pt,leftmargin=*]
\item Assign the smallest suitable shared object to $t$ if it exists.
\item If there are no suitable shared objects, create a new shared object with size $size_t$ and assign it to $t$.
\end{itemize}

\begin{figure}[!t]  % Figure 4.  Page 4 upper left.
  \centering
  \includegraphics[height=4.5cm]{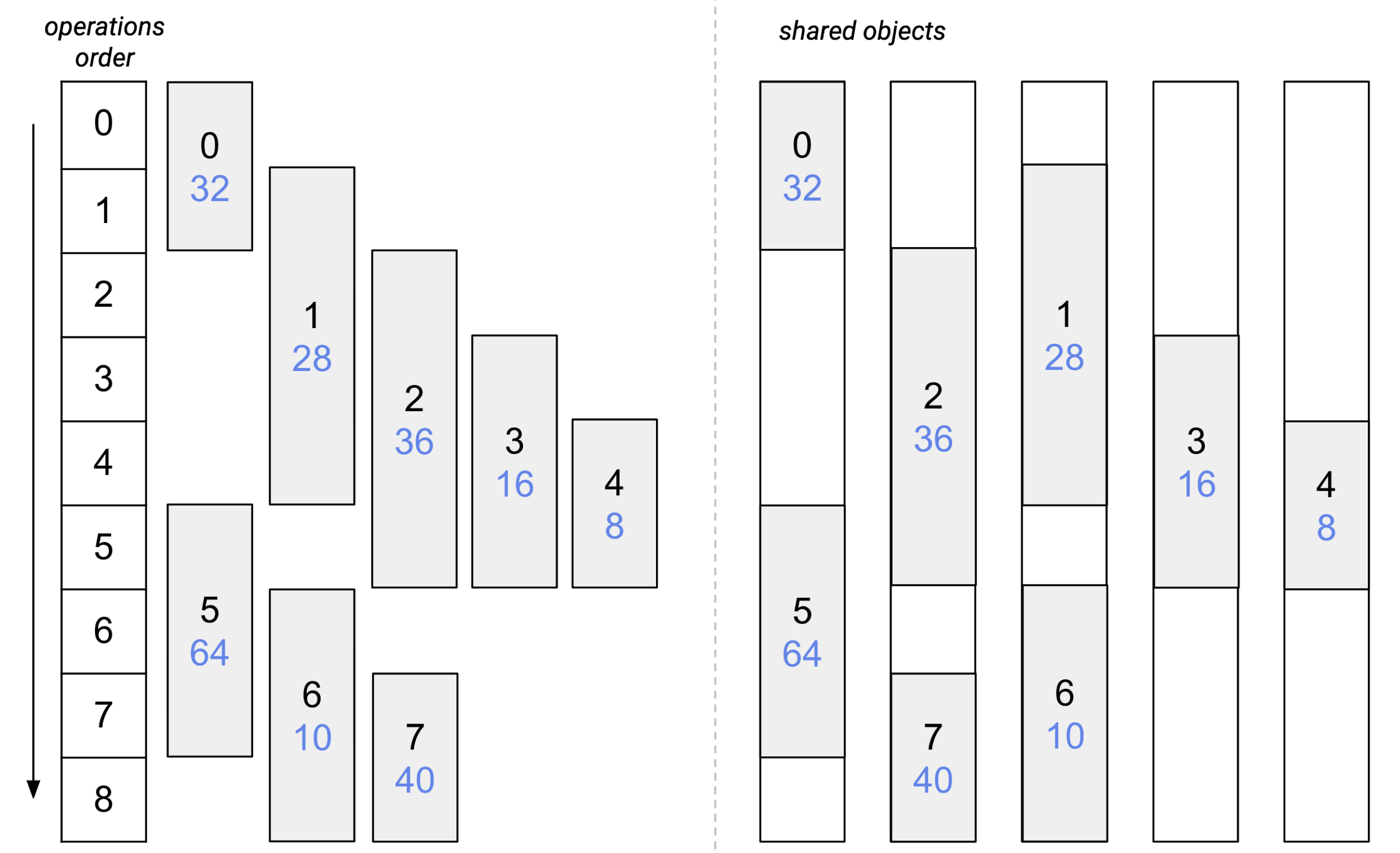}
  \caption{Greedy by Size for Shared Objects on the neural network in Figure~\ref{fig:example}.}
  \label{fig:greedy-by-size}
  \vspace{-0.2cm}
\end{figure}

Greedy by Size has the same complexity as Greedy by Breadth.  Its example output
is shown in Figure~\ref{fig:greedy-by-size}.

\subsection{Greedy by Size Improved}

While refining Greedy by Size, we observed that there were close
mis-assignments that prevented it from reaching the lower bound.  If
there was a wiggle room for tensors with similar sizes, it could have reached
the optimum.
% (a) tensors with similar sizes  have almost equal significance for the result. Second improvement considers, which tensor of those with almost equal significance is better to assign first, and what shared object is the most suitable for such assignment.

As the theoretical lower bound of Shared Objects is determined by positional
maximums, we split Greedy by Size into stages by positional maximum.
In the first stage, we assign all tensors with size equal to
largest positional maximum.  In the second stage, we assign all tensors with
sizes less than the largest positional maximum, but greater than the second
positional maximum.  In the third stage, we assign all tensors with sizes equal
to the second positional maximum, etc.
% In the fourth stage, tensors with sizes
% between the second and the third positional maximums, etc.
until all the tensors
are assigned.  We consider all tensors in one stage to have almost equal
significance.  This is based on the results of the experiments
with greedy algorithms on different neural networks: final result is usually
pretty close to the theoretical lower bound, and most of the shared objects,
especially larger ones, often have the same sizes as in the lower bound.

Another improvement we propose is the order of tensors assignment inside of one
stage.  Tensor sizes in one stage are almost equal, so we choose such a pair of
tensor and shared object that result in the smallest possible time gap when
shared object is not in use, i.e. find such pair of tensor $t$ form current
stage and \textit{suitable} shared object $s$ for it, that distance between
usage interval for $t$ and closest usage interval from tensors, previously
assigned to $s$, is the smallest possible. It means, that we find shared object
that still can be used for tensor $t$ assignment, but the gap when this shared
object is not used right before or right after tensor $t$ usage interval is the
smallest possible.

\begin{figure}[!t]  % Figure 5.  Page 4 upper right.
  \centering
  \includegraphics[height=4.5cm]{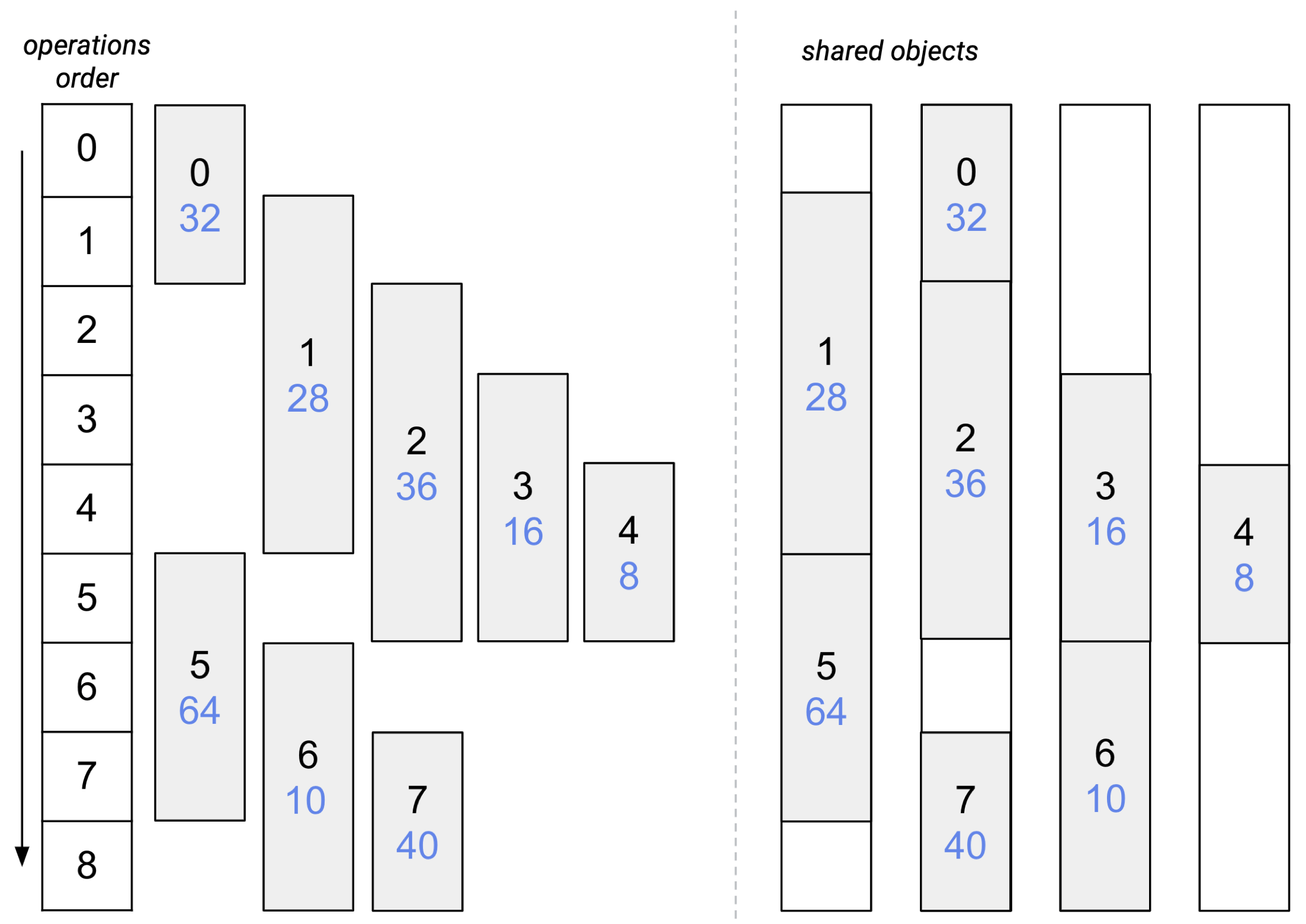}
  \caption{Greedy by Size Improved for Shared Objects on the neural network in Figure~\ref{fig:example}.}
  \label{fig:greedy-by-size-improved}
  \vspace{-0.2cm}
\end{figure}

These improvements can be implemented without changing the complexity of Greedy by Size.
The algorithm assigns shared objects to tensors, using the order defined by positional maximums from ~\ref{fig:profiles}.
Figure~\ref{fig:greedy-by-size-improved} visualizes this strategy.
% Start from tensor 5 which size 64 is equal to the largest positional maximum. Then the algorithm assigning shared object to tensor 7 with size 40 (equal to second largest positional maximum). Sizes of tensors 0, 1 and 2 lie between positional maximums 40 and 16, so they are considered to have almost equal significance. Their processing order is defined by usage intervals distance to previously allocated tensors. Tensor 1 is the closest to tensor 5, and first shared object can be assigned to both of them. After tensor 1 we process tensor 2, because it is pretty close to tensor 7 and can be assigned to the same shared object. After that algorithms assigns the same shared object to remaining tensor 0. Then algorithm proceeds to the next positional maximums in non-increasing order and assigns shared objects to tensors 3, 6 and 4. Resulting sizes of shared objects are 64, 40, 16 and 8, and total memory consumption is the sum of these values, e.g. 64+40+16+8=128. We can note that final Shared Objects assignment in this example is the same as in Figure~\ref{fig:greedy-by-breadth} for Greedy by Breadth algorithm, but of course they are not always the same for different neural networks; check the captions of those Figures for the difference.
The results of experiments confirm that using Greedy by Size Improved provides us with better or the same result, compared to the original without improvements.

\section{The Offset Calculation Approach}

We call the other memory sharing approach \textit{Offset Calculation} where a
large chunk of memory is pre-allocated and the intermediate tensors are given
parts of the memory by the offsets within the memory block.  
The main objective is \textit{to minimize the size of the allocated memory
block}.  While the solution to this problem shows best performance in terms of
total allocated memory, it is only applicable to CPU memory or GPU buffers, but
not GPU textures which need to be accessed as a whole.  The solution of Shared
Objects problem can be converted to the solution of Offset Calculation
problem by placing the shared objects contiguously in memory.  The opposite is
not true as memory footprints of tensors with non-intersecting usage intervals
can still overlap.
% Similar to Section \ref{sec:shared-objects}, we start by investigating the theoretical lower bound.

The Offsets Calculation problem can be seen as a special case of
2D strip packing problem~\cite{sekiyama2018profile}.
A set of rectangular items with fixed coordinates by one axes into a container
to minimize its size by other dimension.  If the height of a container is
treated as the allocation time axis, then we need to minimize the
container's width which corresponds to the memory footprint.

\subsection{Theoretical Lower Bound}

During the execution of any operator of the neural network all tensors in its profile need to be present in memory. Their total size is equal to this operator's breadth. It means, that any strategy will provide us with memory consumption greater or equal to any operator breadth, and the lower bound for Offset Calculation is equal to the maximum among all operator breadths.
The lower bound cannot be always achieved, but our methods achieve the lower bound in most cases.

\subsection{Greedy by Size for Offset Calculation}

As \textit{Greedy by Size} works well for Shared Objects, we employ a
similar method for Offsets Calculation (Algorithm~\ref{alg:greedy-by-size-offsets}).

\begin{algorithm}[!b]
   \begin{small}
   \caption{Greedy by Size for Offset Calculation}
   \label{alg:greedy-by-size-offsets}
\begin{algorithmic}[1]
   \STATE sort tensor~usage~records in non-increasing order of $\mathit{size}$
   \STATE \textbf{for each} $\mathit{t} \in$ tensor~usage~records \textbf{do}
   \STATE \hspace{0.2cm}$\mathit{assigned\_offset_t} \leftarrow \mathit{NIL}$
   \STATE $\mathit{total\_consumption} \leftarrow \mathit{0}$
   \STATE $\mathit{ordered\_allocated\_ids} \leftarrow \emptyset$
   \STATE \textbf{for each} $\mathit{t} \in$ tensor~usage~records \textbf{do}
   \STATE \hspace{0.2cm} $\mathit{prev\_offset} \leftarrow \mathit{0}$
   \STATE \hspace{0.2cm} $\mathit{best\_offset} \leftarrow \mathit{NIL}$
   \STATE \hspace{0.2cm} $\mathit{smallest\_gap} \leftarrow \infty$
   \STATE \hspace{0.2cm} \textbf{for each} $\mathit{x} \in \mathit{ordered\_allocated\_ids}$ \textbf{do}
   \STATE \hspace{0.4cm} $\mathit{max\_first\_op}\leftarrow\max(\mathit{first\_op_t},\mathit{first\_op_x})$
   \STATE \hspace{0.4cm} $\mathit{min\_last\_op}\leftarrow\min(\mathit{last\_op_t},\mathit{last\_op_x})$
   \STATE \hspace{0.4cm} \textbf{if} $\mathit{max\_first\_op}\le\mathit{min\_last\_op}$ \textbf{then}
   \STATE \hspace{0.6cm} $\mathit{gap} \leftarrow \mathit{offset_x} - \mathit{prev\_offset}$
   \STATE \hspace{0.6cm} \textbf{if} $\mathit{gap} \ge \mathit{size_t}$ \textbf{then} $\mathit{gap} < \mathit{smallest\_gap}$ \textbf{then}
   \STATE \hspace{0.8cm} $\mathit{smallest\_gap} \leftarrow \mathit{gap}$
   \STATE \hspace{0.8cm} $\mathit{best\_offset} \leftarrow \mathit{prev\_offset}$
   \STATE \hspace{0.4cm} $\mathit{prev\_offset} \leftarrow \max(\mathit{prev\_offset}, \mathit{offset_x}+\mathit{size_x})$
   \STATE \hspace{0.2cm} \textbf{if} $\mathit{best\_offset} = \mathit{NIL}$ \textbf{then}
   \STATE \hspace{0.4cm} $\mathit{best\_offset} \leftarrow$ $\mathit{prev\_offset}$
   \STATE \hspace{0.2cm} $\mathit{offsets_t} \leftarrow$ $\mathit{best\_offset}$
   \STATE \hspace{0.2cm} $\mathit{total\_consumption}\leftarrow$ \\
          \hspace{0.4cm} $\max(\mathit{total\_consumption}, \mathit{best\_offset}+\mathit{size_t})$
   \STATE \hspace{0.2cm} insert $\mathit{t}$ into $\mathit{ordered\_allocated\_ids}$ % to save non-decreasing order of $first\_op$
\end{algorithmic}
\end{small}
\end{algorithm}

\begin{figure}[!t]  % Figure 6
  \centering
  \includegraphics[height=4.5cm]{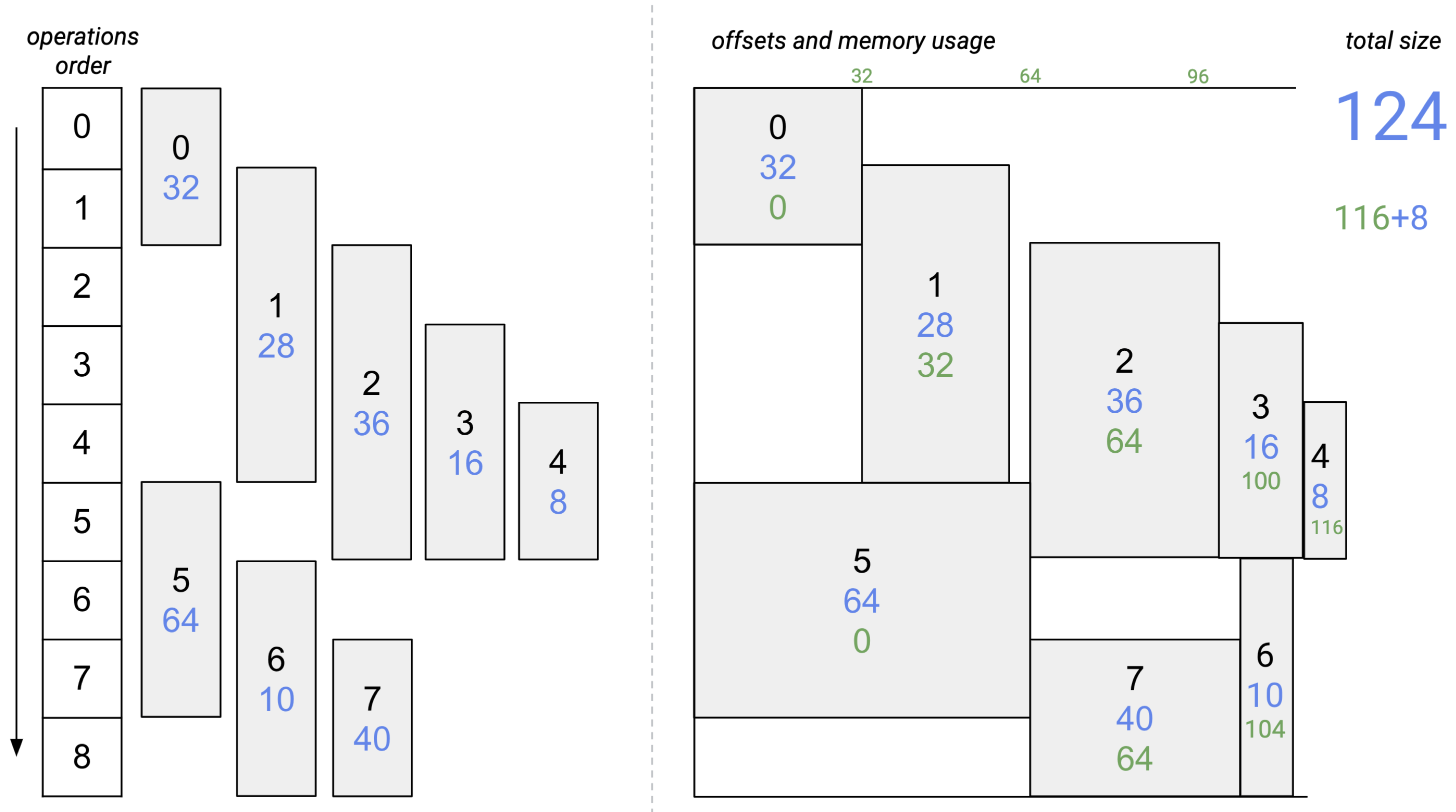}
  \caption{Greedy by Size for Offset Calculation on the neural network in Figure~\ref{fig:example}.
  % Vertical axes is a timeline of operators execution, horizontal axes shows positions of allocated tensors in memory. % Tensor sizes are shown in blue, and resulting offsets are shown in green.}
  }
  \label{fig:greedy-by-size-offsets}
  \vspace{-0.2cm}
\end{figure}

We first iterate through tensor usage records in non-increasing order by their $size$ (L.1,6). For each record, we check already assigned tensors whose usage intervals intersect with that of the current tensor (L.10--13) to find the smallest gap in memory between them such that current tensor fits into that gap (L.9,14--17). If such a gap is found, the current tensor is allocated to this gap.  Otherwise, we allocate it after the rightmost tensor whose usage interval intersect with that of the current tensor (L.19--20).  We assign the corresponding offset to current tensor and the tensor becomes assigned (L.21--23) as shown in Figure~\ref{fig:greedy-by-size-offsets}.  % Example of how this algorithm works for neural network from Figure~\ref{fig:example} can be seen in Figure~\ref{fig:greedy-by-size-offsets}.  %The algorithm calculates offsets for tensors in non-increasing order of tensors size: 5, 7, 2, 0, 1, 3, 6, 4. Resulting consumption is the sum of the rightmost offset and corresponding tensor size, e.g. 116+8=124, which is equal to theoretical lower bound for the result of this approach.

\subsection{Greedy by Breadth for Offset Calculation}

\textit{Greedy by Breadth} can also be converted for Offsets Calculation in a similar fashion.
Specifically:
\begin{itemize}[noitemsep,topsep=0pt,leftmargin=*]
\item Iterate through all operators in non-increasing order by their \textit{breadth}.
\item For each operator in this order, iterate through all tensors from its profile that have not been assigned yet, in non-increasing order of their \textit{size}.
\item To calculate the offset for the tensor, use the same logic of finding the smallest gap as in Alg.~\ref{alg:greedy-by-size-offsets} (L.7--23).
At the end of this step, the tensor is marked as assigned.
\end{itemize}

% As we have seen from experiments (Table~\ref{offset-calc-results}), 
While Greedy by Breadth does not perform well for Offset Calculation compared to
Greedy by Size, % as discussed in Section~\ref{sec:eval},
it still outperforms the prior work on some networks, e.g.~MobileNet v2.

\begin{table*}[!t]  % Table 1
\caption{Memory footprint of Shared Objects strategies (in MB; best results in
bold).  The first 3 rows are our strategies, the next 2 are prior work, and the
last 2 serve as baseline.}
\label{shared-objects-results}
\begin{center}
\begin{small}
\begin{tabular}{lcccccc}
\toprule
Strategy & MobileNet v1 & MobileNet v2 & DeepLab v3 & Inception v3 & PoseNet & BlazeFace \\
\midrule
Greedy by Size    & \textbf{4.594} &	7.178 &	\textbf{6.437} & \textbf{10.337} & \textbf{6.347} & 0.592 \\
Greedy by Size Improved & \textbf{4.594} & 6.891 & \textbf{6.437} & \textbf{10.337} & \textbf{6.347} & \textbf{0.518} \\
Greedy by Breadth & 6.125 &	\textbf{6.699} & \textbf{6.437}  & 10.676 & 8.390 & 0.675 \\
\midrule
Greedy~\cite{lee2019device}     & \textbf{4.594} & 8.039 & 7.168 & 12.703 & \textbf{6.347} & 0.587 \\
Min-cost Flow~\cite{lee2019device}      & 5.359 & 7.513 & 8.364 & 10.624  & 7.359 & 0.582 \\
\midrule
Lower Bound & 4.594 & 6.604 & 6.105 & 8.955 & 6.347 & 0.518 \\
Na\"ive & 19.248 & 26.313 & 48.642 & 54.010 & 28.556 & 2.698 \\
\bottomrule
\end{tabular}
\end{small}
\end{center}
\end{table*}

\begin{table*}[!t]  % Table 2
\caption{Memory footprint of Offset Calculation strategies (in MB; best results
in bold).  The first 2 rows are our strategies, the next 2 are prior work, and
the last 2 serve as baseline.}
\label{offset-calc-results}
\begin{center}
\begin{small}
\begin{tabular}{lcccccc}
\toprule
Strategy & MobileNet v1 & MobileNet v2 & DeepLab v3 & Inception v3 & PoseNet & BlazeFace \\
\midrule
Greedy by Size & \textbf{4.594} & \textbf{5.742} & 4.653 & \textbf{7.914} & \textbf{6.271} & \textbf{0.492} \\
Greedy by Breadth & \textbf{4.594} & \textbf{5.742} & 4.653 & \textbf{7.914} & 7.359 & 0.656 \\
\midrule
Greedy~\cite{lee2019device} & 6.125 & 6.508 & 4.985 & 10.624 & 8.362 & \textbf{0.492} \\
Strip Packing~\cite{sekiyama2018profile} & \textbf{4.594} & 6.029 & \textbf{4.321} & \textbf{7.914} & \textbf{6.271} & 0.533 \\
\midrule
Lower Bound & 4.594 & 5.742 & 4.320 & 7.914 & 6.271 & 0.492 \\
Na\"ive & 19.248 & 26.313 & 48.642 & 54.010 & 28.556 & 2.698 \\
\bottomrule
\end{tabular}
\end{small}
\end{center}
\end{table*}

\section{Evaluation}
\label{sec:eval}

We compare our strategies with Greedy~\cite{lee2019device},
Min-cost Flow~\cite{lee2019device},
and Strip Packing Bestfit~\cite{sekiyama2018profile} on
MobileNet v1~\cite{howard2017mobilenets},
MobileNet v2~\cite{sandler2018mobilenetv2},
DeepLab v3~\cite{chen2017deeplab},
Inception v3~\cite{szegedy2016rethinking},
PoseNet~\cite{kendall2015posenet},
and BlazeFace~\cite{bazarevsky2019blazeface},
at 32-bit precision floating point,
but the strategies can be generalized to any data type.

The best results for Shared Objects are achieved with Greedy by Size Improved
on all networks except MobileNet v2, for which Greedy by Breadth does better
(Table~\ref{shared-objects-results}).  Compared to prior work, our strategies do
up to 11\% better, and compared to na\"ive strategy, they do up to 7.5$\times$
better.  The most significant improvement is seen for DeepLab with all three
described strategies, for MobileNet v2 with Greedy by Breadth, and for BlazeFace
with Greedy by Size Improved. Our strategies reach the theoretical lower
bound for MobileNet v1, PoseNet, and BlazeFace, and are within 16\% of the lower
bound for the other networks.  For inference engines needing the Shared Objects
approach, it is recommended to default to Greedy by Size Improved.

For Offset Calculation, Greedy by Size performs best as shown in Table
\ref{offset-calc-results}.  It achieves the theoretical lower bound on all
selected neural networks, except DeepLab, where it still falls within 8\% of
the lower bound.  Moreover, it provides us with memory allocation consuming up
to 25\% less memory for intermediate tensors than Greedy,
up to 7.7\% less than in Strip Packing Bestfit, and up to 10.5 times less than in a
na\"ive strategy.
Only for DeepLab, Strip Packing Bestfit shows 7.2\% better allocation that
is very close to the theoretical lower bound.  For inference engines requiring
the Offset Calculation approach, it is recommended to evaluate both Greedy by
Size and Strip Packing Bestfit before the first inference and select the
superior performing strategy.

\section{Conclusion}

We presented five novel strategies for efficiently sharing memory buffers among
intermediate tensors in deep neural networks to minimize the memory footprint
of the inference engine at the edge.  The experiments showed that our strategies
get the inference run-time's memory footprint to equal to or close to the
theoretical lower bound.

The presented strategies for either approach are fast enough (a few milliseconds for most
networks), so that they can be explored at run-time for the smallest memory
footprint.  In general, i.e. CPU inference or GPU inference with buffers, it is
recommended to explore the two best strategies for the Offset Calculation problem,
Greedy by Size and Strip Packing Bestfit.  For the Shared Objects problem, e.g.~GPU
inference with textures, Greedy by Size Improved and Greedy by Breadth are
recommended for pre-inference exploration.

The strategies presented assume that the sizes of intermediate tensors are known
in advance.  This assumption may not be true for recurrent neural networks with
long short-term memory units~\cite{hochreiter1997long} including intermediate
tensors with dynamically changing sizes.  For such cases, the algorithms need to
be run multiple times saving information about allocation from all runs in one
place.  The first run will allocate only those tensors whose sizes are known at
the beginning, and the second run will allocate those tensors whose sizes become
known after calculation of the first dynamic tensor, etc.

\subsection{Future Work}

The operator indices in tensor usage records and intervals are defined by the
topological sort of the neural network.  Optimizing the sorting algorithm for
the smallest possible memory footprint is a potential future research topic.

The current choice for the best strategy only focuses on the memory footprint.
Other criteria such as cache hit rate and inference latency can be incorporated
into evaluation for fast inference on resource-constrained systems.

\section*{Acknowledgements}
We would like to thank Andrei Kulik for the initial brainstorming and the TFLite team
for adopting our strategies to TFLite's memory manager, especially Terry Heo for
the final implementation.

\bibliography{main}
\bibliographystyle{sysml2019}

\end{document}